\documentclass{article}

\usepackage{microtype}
\usepackage{graphicx}
\usepackage{subfigure}
\usepackage{booktabs} 
\usepackage{tablefootnote}
\usepackage{float}

\usepackage{hyperref}

\usepackage[accepted]{icml2021}

\icmltitlerunning{Domain Adaptation for Low-Resource GIS ML}

\begin{document}


\twocolumn[
\icmltitle{Leveraging Domain Adaptation for \\
           Low-Resource Geospatial Machine Learning}



\icmlsetsymbol{equal}{*}

\begin{icmlauthorlist}
\icmlauthor{Jack Lynch}{to,mas}
\icmlauthor{Sam Wookey}{mas}
\end{icmlauthorlist}

\icmlaffiliation{to}{Department of Electrical Engineering, NC State University, Raleigh, North Carolina, USA}
\icmlaffiliation{mas}{Masterful.AI, San Francisco, California, USA}

\icmlcorrespondingauthor{Jack Lynch}{jmlynch3@ncsu.edu, jack@masterful.ai}

\icmlkeywords{Machine Learning, ICML, GIS, Domain Adaptation}

\vskip 0.3in
]



\printAffiliationsAndNotice{}  

\begin{abstract}

Machine learning in remote sensing has matured alongside a proliferation in availability and resolution of geospatial imagery, but its utility is bottlenecked by the need for labeled data. What's more, many labeled geospatial datasets are specific to certain regions, instruments, or extreme weather events. We investigate the application of modern domain-adaptation to multiple proposed geospatial benchmarks, uncovering unique challenges and proposing solutions to them.

\end{abstract}

\section{Introduction}
\label{introduction}

The use of machine learning for remote sensing has matured alongside an increase in the availability and resolution of satellite imagery, enabling advances in such tasks as land use classification \cite{Campos-Taberner2020}, natural risk estimation \cite{Risk}, disaster damage assessment \cite{xu2019building}, and agricultural forecasting \cite{ag}.

However, many labeled geospatial datasets---including all of those used in the previously cited works---only cover \emph{specific regions} of the world, limiting their utility and adoption elsewhere. Xu et al. explicitly cite this limitation as a focus of future work. Furthermore, these datasets are often limited to a single instrument and method of acquisition, limiting their generalization to acquisitions made by other instruments and at other angles.

Domain adaptation attempts to leverage labeled ``source'' data toward learning on some separate ``target'' dataset, for which only unlabeled data exists. In the context of climate-focused machine learning, it provides a potential framework for scaffolding innovations beyond their limiting datasets to other areas, instruments, and contexts. 

In this proposal, we claim domain adaptation will be necessary to widen the equitable adoption of machine learning for geospatial applications, enabling its use for environmental monitoring and forecasting in areas of the world without labeled data. We propose multiple geospatial domain-adaptation benchmarks, explore the challenges they pose to traditional domain-adaptation algorithms, and suggest methods of improvement.

\section{Geospatial Domain Adaptation}

The application of domain adaptation to geospatial machine learning requires both datasets and methods tailored to the context of remote sensing.

\subsection{Proposed Benchmarks}


For our initial experiments we propose two domain-adaptation benchmarks derived from the well-known SpaceNet datasets of labeled satellite imagery \cite{vanetten2019spacenet}. As noted by \citet{mrv}, land use determines a significant portion of human carbon emission; as such, we focus on building segmentation as a useful tool in the monitoring and prediction of emissions. We measure the Jaccard index \cite{jaccard}, or intersection over union (IoU), of target-dataset segmentations as our primary performance metric. 

SpaceNet-2 contains imagery and corresponding building annotations for four cities: Las Vegas, Paris, Shanghai, and Khartoum. We may treat one subset of cities as the ``source'' and another subset as the ``target'' and attempt adaptation from one to the other: for example, from Las Vegas to Khartoum. The ability to generalize building segmentation from well-labeled cities to growing ones in different areas of the world would greatly increase the utility of such methods. Given compute constraints, we restrict our experiments here to those treating Khartoum as the target dataset.

SpaceNet-4 contains imagery of specific regions for varying angles off-nadir, loosely grouped into ``on-nadir'' for low angles, ``off nadir'' for moderate angles, and ``very off-nadir'' for large angles. Off-nadir imagery introduces a lower effective resolution and is marked in urban settings by building ``tilt'' relative to on-nadir imagery. We can attempt adaptation from on-nadir to off-nadir or very off-nadir imagery, from off-nadir to very off-nadir, etc. As off-nadir imagery is often the quickest path to imaging during disasters \cite{offnadir}, the ability to extend more-common ``on-nadir'' performance to off-nadir scenarios could aid in accelerating disaster response.

\begin{figure*}
\centering     
\subfigure[Las Vegas]{\label{fig:vegas}\includegraphics[width=40mm]{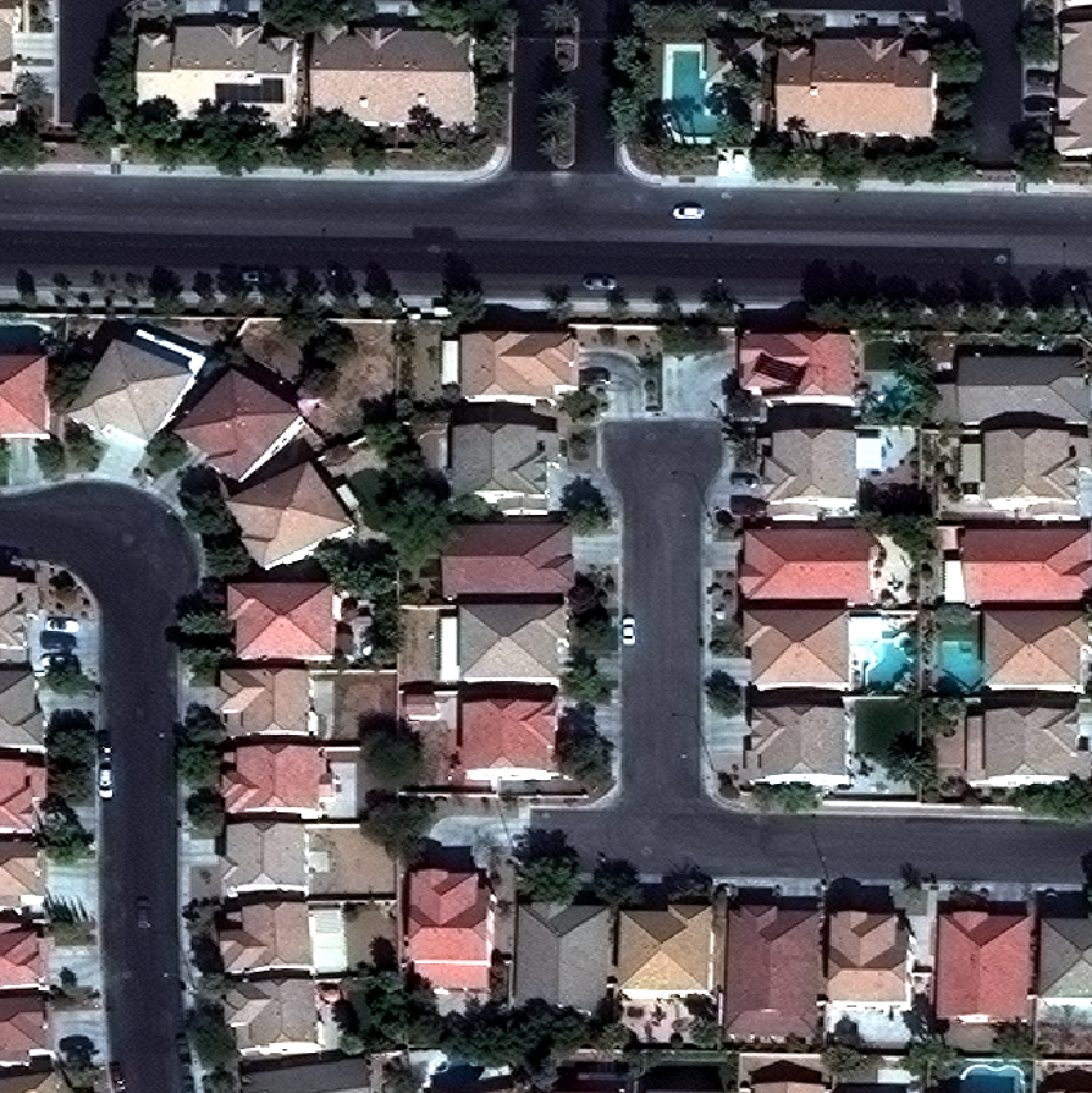}}
\subfigure[Khartoum]{\label{fig:khartoum}\includegraphics[width=40mm]{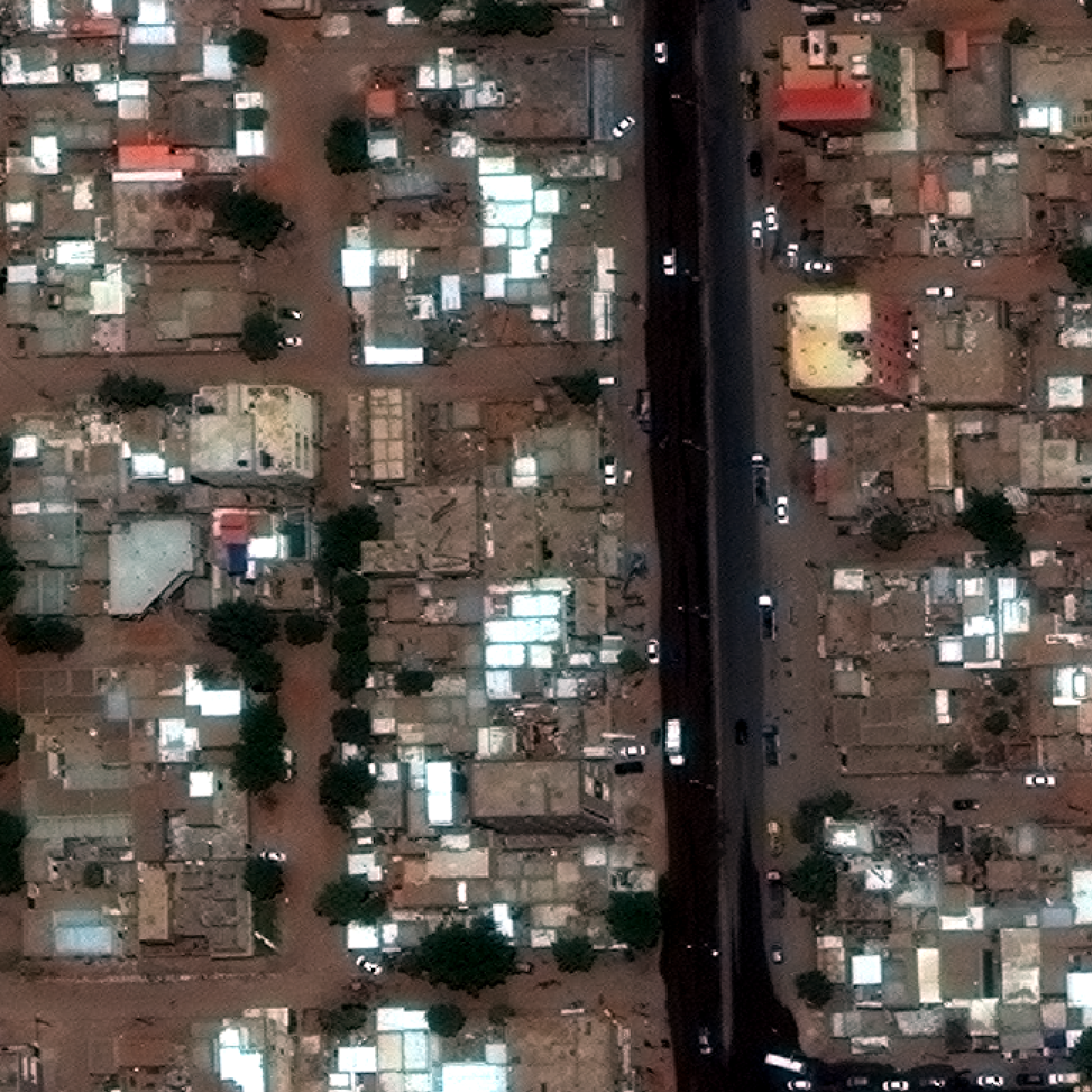}}
\subfigure[On-Nadir]{\label{fig:onnadir}\includegraphics[width=40mm]{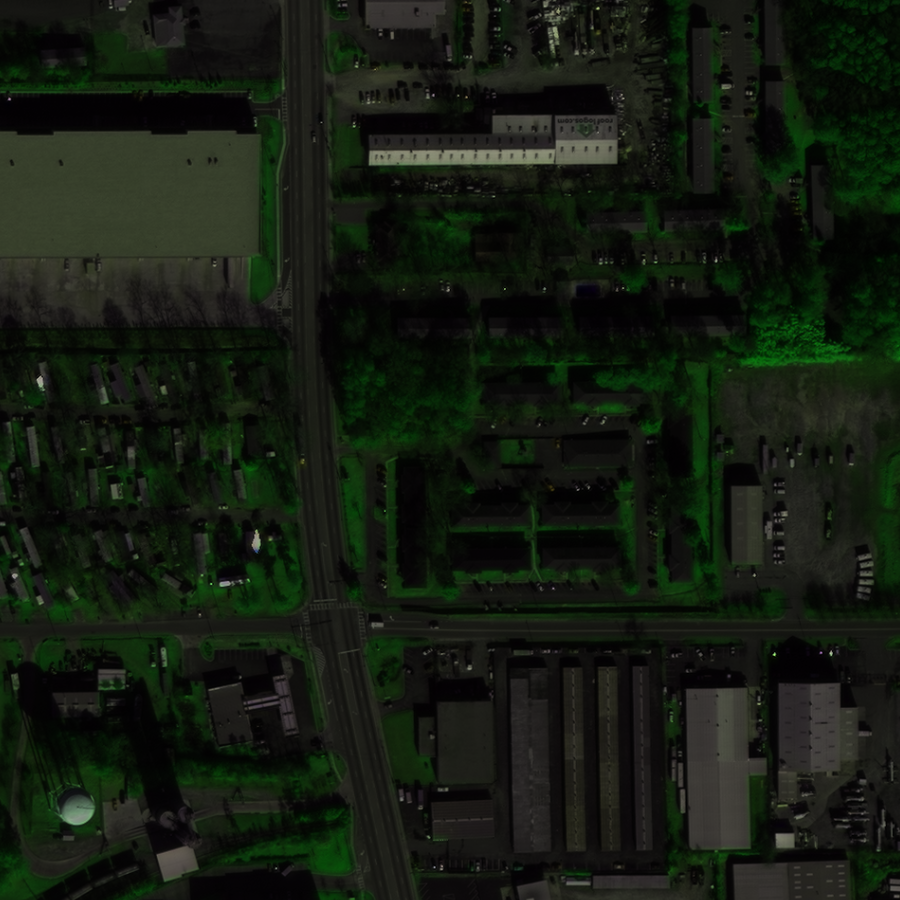}}
\subfigure[Very Off-Nadir]{\label{fig:voffnadir}\includegraphics[width=40mm]{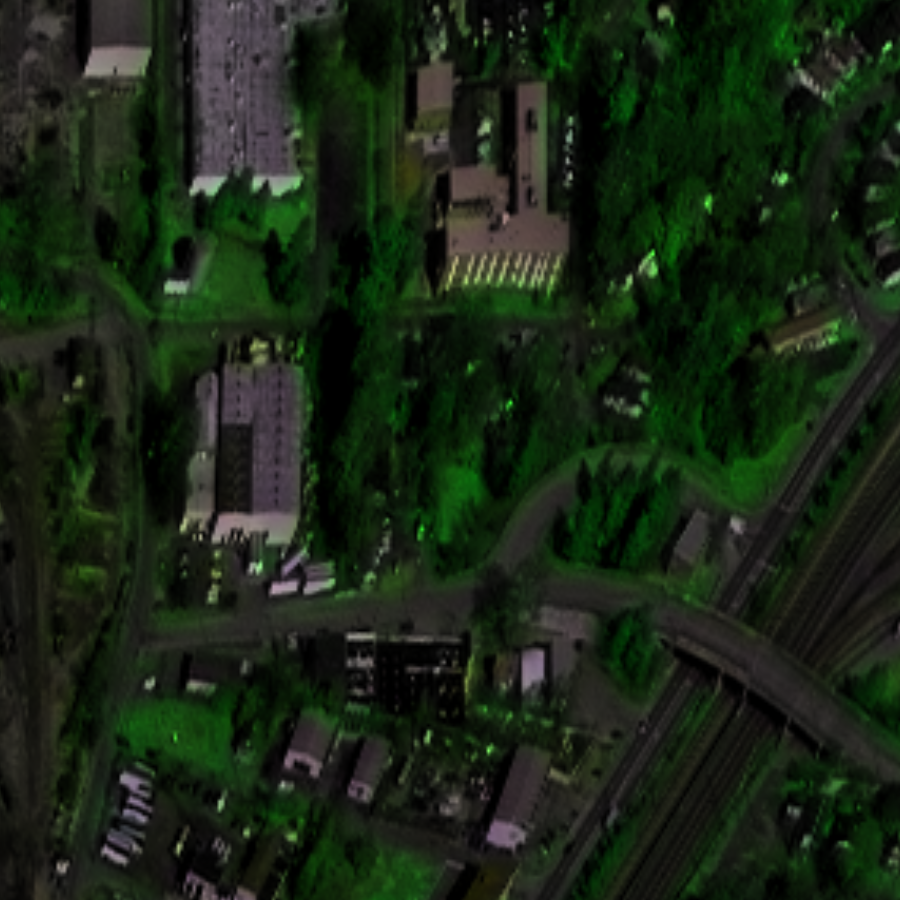}}
\caption{Example acquisitions from the SpaceNet-2 (left) and SpaceNet-4 (right) datasets. The shift from Las Vegas to Khartoum is characterized by different colors, textures, building styles and arrangements, and (occasionally) the angle off-nadir. When moving from On- to Off-Nadir data, buildings move from ``top-down'' footprints to three-dimensional objects, and the effective resolution degrades.}
\label{spacenet}
\end{figure*}

Samples from these datasets are shown in Figure~\ref{spacenet}. Notably, off-nadir acquisition angle varies significantly between cities in SpaceNet-2, suggesting the two benchmarks feature some overlap in sources of difficulty. 

\subsection{Methods}


For our experiments we adopt the ADVENT approach of \citet{vu2019advent}, a relatively simple adversarial domain adaptation method whose precursor \cite{tsai2020learning} serves as a first stage for the current state of the art \cite{zhang2021prototypical}. We perform all experiments using the same hyperparameters provided in \citet{vu2019advent}.

Notably, the ADVENT method, alongside precursors like \citet{tsai2020learning, hoffman2017cycada}, assumes that source and target label distributions are significantly more similar than their corresponding input distributions; otherwise, the discriminator would be able to distinguish between even exceptionally domain-general source and target predictions, diminishing its utility as a regularizer. 

\subsection{Challenges}

\subsubsection{Limited Source Data}


Most common domain-adaptation benchmarks---for example, VisDA \cite{peng2017visda} in classification or GTA-to-CityScapes \cite{richter2016playing, Cordts2016Cityscapes} in segmentation---provide a large amount of labeled source data; as such, many recent domain-adaptation algorithms have not been examined in the context of limited source data (what could be considered ``few-shot'' domain adaptation).

The few-shot limitations specific to adversarial domain adaptation are not well understood, but they are especially relevant to geospatial machine learning, where new unlabeled data appears daily but labeled datasets are often small. For example, in our city-to-city benchmarks, no single city features more than $5,000$ labeled samples, while none of on-nadir, off-nadir, or very-off-nadir collections feature more than $10,000$.

\begin{figure*}
\centering     
\subfigure[Input]{\label{fig:vegas}\includegraphics[width=40mm]{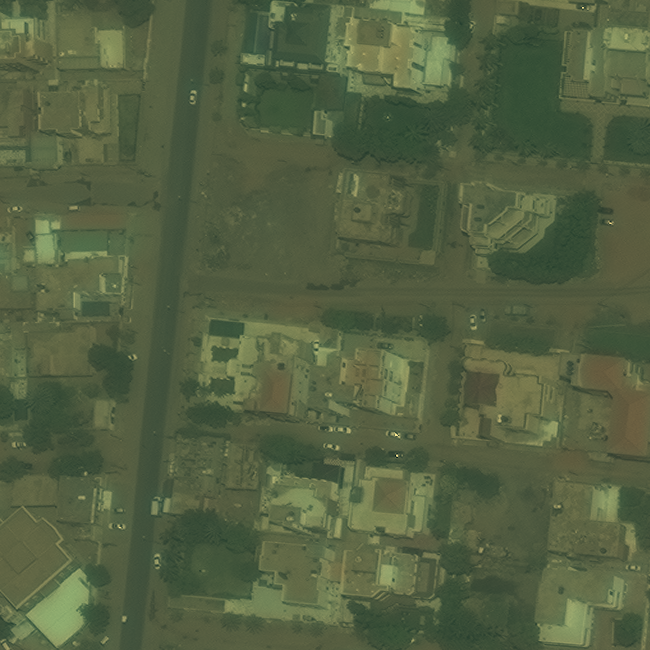}}
\subfigure[Ground Truth]{\label{fig:khartoum}\includegraphics[width=40mm]{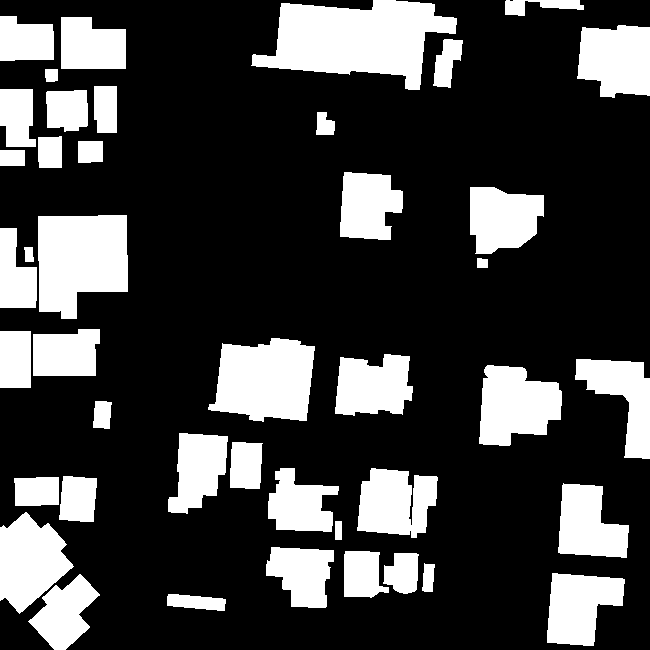}}
\subfigure[ADVENT]{\label{fig:onnadir}\includegraphics[width=40mm]{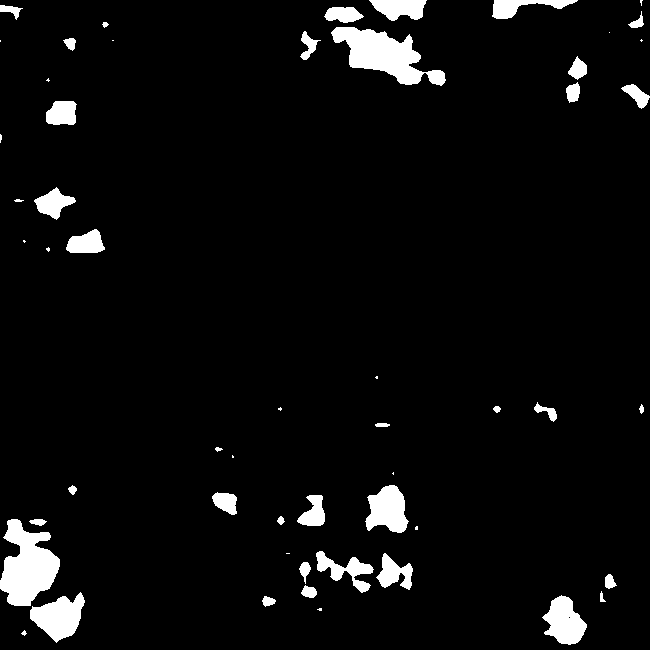}}
\subfigure[Source-Only]{\label{fig:voffnadir}\includegraphics[width=40mm]{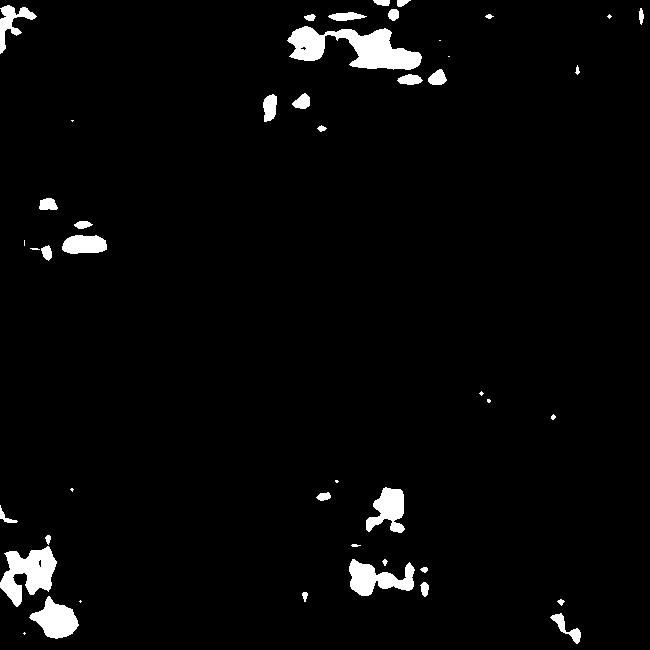}}
\caption{Example predictions from both ADVENT and source-only (non-adaptive) Vegas-to-Khartoum training for (a). Both ADVENT and source-only approaches fail to identify most ``building'' pixels, and ADVENT does not improve over source-only.}
\label{preds}

\end{figure*}

We observe that this limited source data hampers the adaptive ability of ADVENT, often performing worse than source-only training as shown in Table~\ref{spacenet-tab} and Figure~\ref{preds}\footnote{For GTA-to-CityScapes, we report the ADVENT metric from \citet{vu2019advent} and the source-only metric from \citet{tsai2020learning}. Though we didn't replicate the results of Vu et al. due to compute constraints, we extended their open-sourced code for our own experiments.}. Though ADVENT trains stably for hundreds of thousands of iterations on traditional domain-adaptation benchmarks, on most city-to-city and nadir-angle paired datasets it can only train for thousands before beginning to diverge. We attribute this poor performance to discriminator overfitting \cite{ada}, leading to the injection of unhelpful gradients during training.

\begin{table*}[t]
\caption{Improvement in target Intersection over Union (IoU) using ADVENT, relative to source-only training. V, S, P, and K represent SpaceNet-2 data for cities Las Vegas, Shanghai, Paris, and Khartoum, respectively. On and V. Off represent SpaceNet-4 ``on-nadir'' and ``very off-nadir'' subsets, respectively. GTA and CS represent the common domain-adaptation benchmark datasets of Grand Theft Auto and CityScapes, respectively. Certain combinations were omitted due to compute constraints.}
\label{spacenet-tab}
\vskip 0.15in
\begin{center}
\begin{small}
\begin{sc}
\begin{tabular}{lcccccc}
\toprule
& GTA $\rightarrow$ CS & V $\rightarrow$ K & V, P $\rightarrow$ K & P, S $\rightarrow$ K & V, S, P $\rightarrow$ K & On $\rightarrow$ V. Off \\
\midrule
IoU (ADVENT)    & 47.6 &  13.59 & 9.95 & 26.36 & 25.05& 11.03\\
IoU (src-only)    & 36.6  &  15.09 & 17.56 & 23.62 & 30.09 & 14.77 \\
\midrule
$\Delta$ IoU    & $+11.0$ &  $-1.50$ & $-7.61$ & $+2.74$ & $-5.04$ & $-3.74$ \\
\bottomrule
\end{tabular}
\end{sc}
\end{small}
\end{center}
\vskip -0.1in
\end{table*}

\subsubsection{Disparate Label Distributions}

As mentioned previously, adversarial methods like ADVENT and its predecessors assume that source and target label distributions are significantly more similar than their corresponding image/input distributions. When label distributions are dissimilar, source and target predictions can be more easily distinguished at the same level of domain-generality. 

Visual inspection suggests that the cities in SpaceNet-2 feature larger gaps between their label distributions than do GTA and CityScapes (Figure~\ref{labz}, in the appendix), but this remains to be rigorously examined. We plan to consult structure-preserving dimensionality reduction through methods like UMAP \cite{mcinnes2020umap, sainburg2021parametric} to assess the relative ``gaps'' between pairs of source and target label distributions.

More specifically, a relative, quantitative assessment could be made by applying persistent homology to UMAP complexes of mixed source and target labels, observing average cluster ``label'' purity (here, referring to the binary distinction of source versus target) as a function of increasing similarity. Integrals of these purity curves---one for every pair of source/target label distributions---could be used to compare cluster separability between pairs of label distributions, in turn suggesting their inherent ease of discrimination during adversarial training. 

\begin{figure*}

\centering     
\subfigure[GTAV]{\label{fig:vegas}\includegraphics[width=47mm]{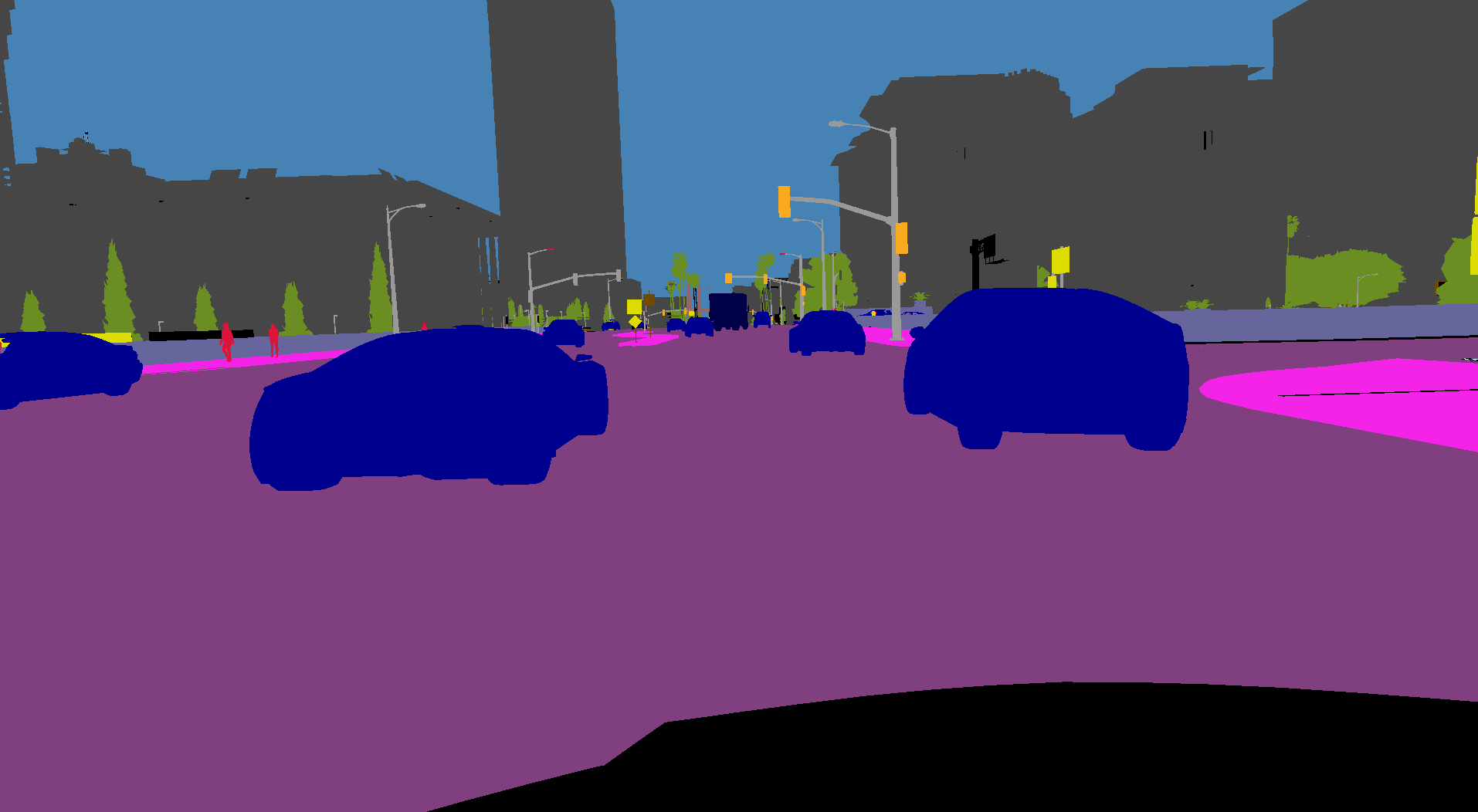}}
\subfigure[CityScapes]{\label{fig:khartoum}\includegraphics[width=47mm]{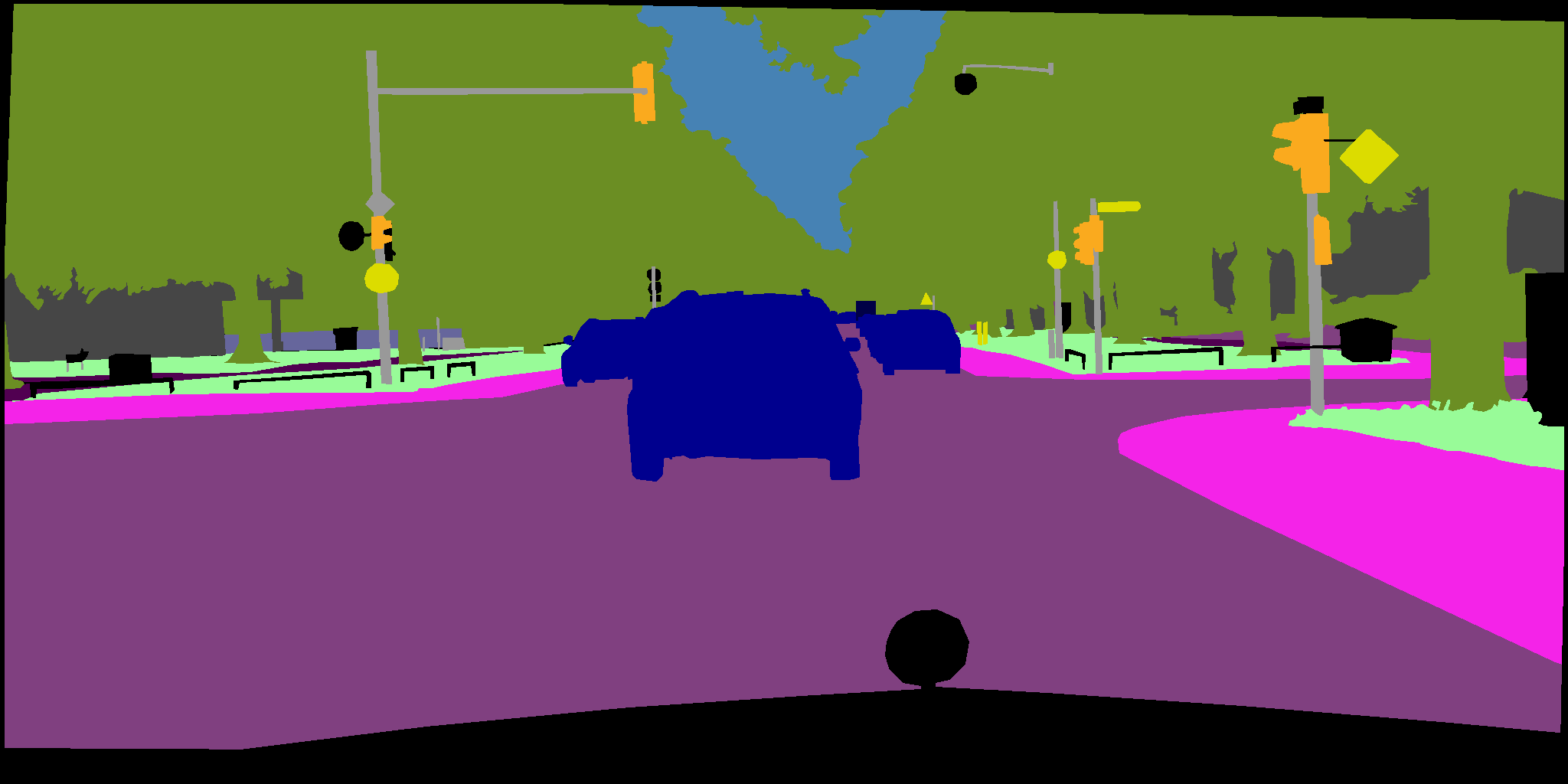}}
\subfigure[Las Vegas]{\label{fig:onnadir}\includegraphics[width=33mm]{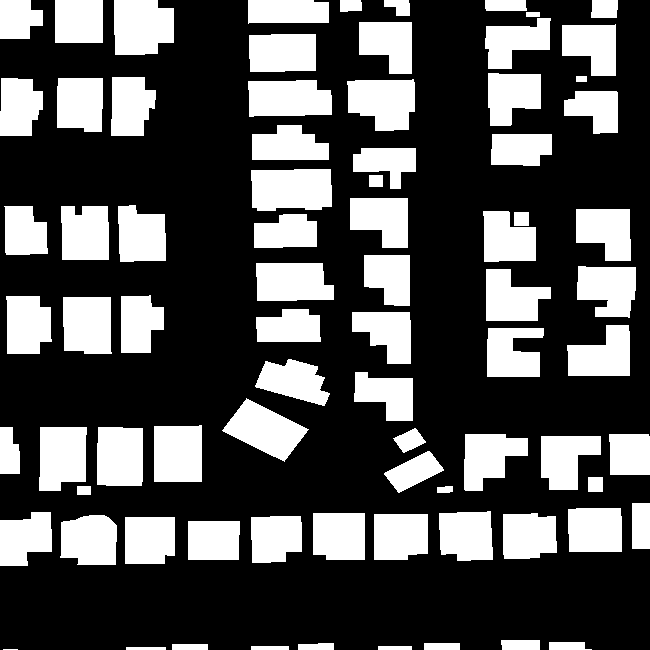}}
\subfigure[Khartoum]{\label{fig:voffnadir}\includegraphics[width=33mm]{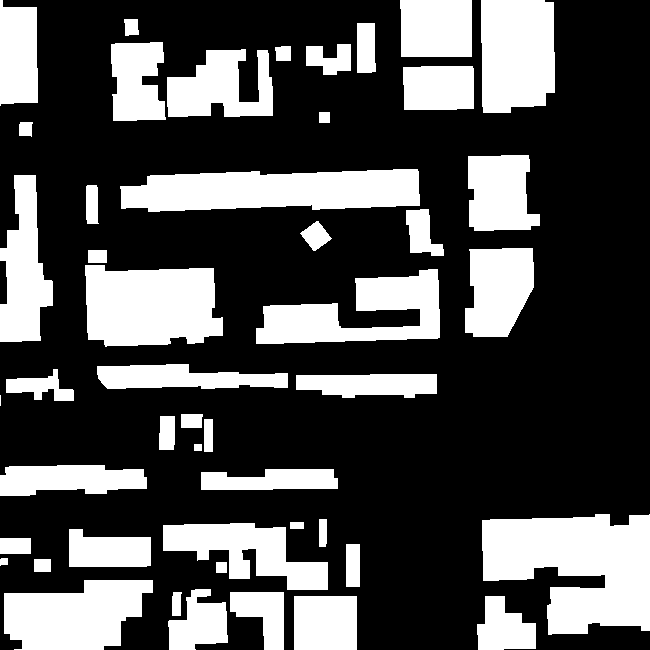}}
\caption{Example labels for GTA, CityScapes, and the Las Vegas and Khartoum subsets of SpaceNet-2. Though GTA features some finer-grain labeling detail, it might otherwise be difficult to tell its labels apart from those of CityScapes. In contrast, the binary labels of Las Vegas and Khartoum are easily distinguished, most notably by the difference in building shapes and arrangement.}
\label{labz}
\end{figure*}

\section{Proposed Solutions}

To address the problems described above, we propose the use of an adaptive discriminator as introduced in \citet{ada}. Though our needs differ from theirs---we do not fear augmentation leakage---an adaptive discriminator should address both issues encountered:
\begin{itemize}
\item It has been shown effective at avoiding discriminator overfitting in low-data regimes; and
\item The geometric class of adaptive augmentations (rotations, affine transforms, etc.) can improve overlap in label distributions, potentially further improving adversarial performance.
\end{itemize}

Early results from the introduction of aggressive augmentations are promising (see Appendix).

With respect specifically to adaptation across instruments or collection methods, as with on-to-off-nadir, we will investigate modeling instrument parameters (e.g. off-nadir angle) through matrix capsules \cite{e2018matrix}, adapted to segmentation per \citet{segcaps}. 

There is also work to be done in the construction of more targeted geospatial domain-adaptation benchmark datasets. For example, as off-nadir imagery is commonly used in disaster response, a dataset for on-to-off-nadir damage estimation would be more directly related to near-term applications.

Finally, geospatial domain adaptation has the unique facet of near-unlimited unlabeled data; as such, in situations where an appropriately delimited target dataset can be automatically aggregated, domain-adaptive methods with limited source data might benefit greatly from self-supervised pretraining, e.g. with \citet{barlow}.


\section*{Acknowledgements}

The authors would like to thank Leland McInnes for his direction in determining how best to apply structure-preserving dimensionality reduction to comparisons of distribution similarity, as well as \citet{vu2019advent} and \citet{ada} for their open-source code, which we extended. Finally, the authors would like to thank our anonymous reviewers for their feedback and Kartik Chandra for his suggestions while revising.

\bibliography{example_paper}

\begin{thebibliography}{21}
\providecommand{\natexlab}[1]{#1}
\providecommand{\url}[1]{\texttt{#1}}
\expandafter\ifx\csname urlstyle\endcsname\relax
  \providecommand{\doi}[1]{doi: #1}\else
  \providecommand{\doi}{doi: \begingroup \urlstyle{rm}\Url}\fi

\bibitem[Campos-Taberner et~al.(2020)Campos-Taberner, Garc{\'i}a-Haro,
  Mart{\'i}nez, Izquierdo-Verdiguier, Atzberger, Camps-Valls, and
  Gilabert]{Campos-Taberner2020}
Campos-Taberner, M., Garc{\'i}a-Haro, F.~J., Mart{\'i}nez, B.,
  Izquierdo-Verdiguier, E., Atzberger, C., Camps-Valls, G., and Gilabert, M.~A.
\newblock Understanding deep learning in land use classification based on
  sentinel-2 time series.
\newblock \emph{Scientific Reports}, 10\penalty0 (1):\penalty0 17188, Oct 2020.
\newblock ISSN 2045-2322.
\newblock \doi{10.1038/s41598-020-74215-5}.
\newblock URL \url{https://doi.org/10.1038/s41598-020-74215-5}.

\bibitem[Cerr{\'o}n et~al.(2020)Cerr{\'o}n, Bazan, and Coronado]{Risk}
Cerr{\'o}n, B., Bazan, C., and Coronado, A.
\newblock {Detection of housing and agriculture areas on dry-riverbeds for the
  evaluation of risk by landslides using low-resolution satellite imagery based
  on deep learning. Study zone: Lima, Peru}.
\newblock In \emph{ICML 2020 Workshop: Tackling Climate Change with Machine
  Learning}, 2020.

\bibitem[Cordts et~al.(2016)Cordts, Omran, Ramos, Rehfeld, Enzweiler, Benenson,
  Franke, Roth, and Schiele]{Cordts2016Cityscapes}
Cordts, M., Omran, M., Ramos, S., Rehfeld, T., Enzweiler, M., Benenson, R.,
  Franke, U., Roth, S., and Schiele, B.
\newblock The cityscapes dataset for semantic urban scene understanding.
\newblock In \emph{Proc. of the IEEE Conference on Computer Vision and Pattern
  Recognition (CVPR)}, 2016.

\bibitem[Dao et~al.(2019)Dao, Rausch, and Zhang]{mrv}
Dao, D., Rausch, J., and Zhang, C.
\newblock {GeoLabels: Towards Efficient Ecosystem Monitoring using Data
  Programming on Geospatial Information}.
\newblock In \emph{NeurIPS 2019 Workshop: Tackling Climate Change with Machine
  Learning}, 2019.

\bibitem[Etten et~al.(2019)Etten, Lindenbaum, and
  Bacastow]{vanetten2019spacenet}
Etten, A.~V., Lindenbaum, D., and Bacastow, T.~M.
\newblock Spacenet: A remote sensing dataset and challenge series, 2019.

\bibitem[Hinton et~al.(2018)Hinton, Sabour, and Frosst]{e2018matrix}
Hinton, G.~E., Sabour, S., and Frosst, N.
\newblock Matrix capsules with {EM} routing.
\newblock In \emph{International Conference on Learning Representations}, 2018.
\newblock URL \url{https://openreview.net/forum?id=HJWLfGWRb}.

\bibitem[Hoffman et~al.(2017)Hoffman, Tzeng, Park, Zhu, Isola, Saenko, Efros,
  and Darrell]{hoffman2017cycada}
Hoffman, J., Tzeng, E., Park, T., Zhu, J.-Y., Isola, P., Saenko, K., Efros,
  A.~A., and Darrell, T.
\newblock Cycada: Cycle-consistent adversarial domain adaptation, 2017.

\bibitem[Jaccard(1912)]{jaccard}
Jaccard, P.
\newblock The distribution of the flora in the alpine zone.1.
\newblock \emph{New Phytologist}, 11\penalty0 (2):\penalty0 37--50, 1912.
\newblock \doi{https://doi.org/10.1111/j.1469-8137.1912.tb05611.x}.
\newblock URL
  \url{https://nph.onlinelibrary.wiley.com/doi/abs/10.1111/j.1469-8137.1912.tb05611.x}.

\bibitem[Karras et~al.(2020)Karras, Aittala, Hellsten, Laine, Lehtinen, and
  Aila]{ada}
Karras, T., Aittala, M., Hellsten, J., Laine, S., Lehtinen, J., and Aila, T.
\newblock Training generative adversarial networks with limited data, 2020.

\bibitem[LaLonde \& Bagci(2018)LaLonde and Bagci]{segcaps}
LaLonde, R. and Bagci, U.
\newblock Capsules for object segmentation, 2018.

\bibitem[Lees et~al.(2020)Lees, Tseng, Dadson, Hern{\'a}ndez, G.~Atzberger, and
  Reece]{ag}
Lees, T., Tseng, G., Dadson, S., Hern{\'a}ndez, A., G.~Atzberger, C., and
  Reece, S.
\newblock {A Machine Learning Pipeline to Predict Vegetation Health}.
\newblock In \emph{ICML 2020 Workshop: Tackling Climate Change with Machine
  Learning}, 2020.

\bibitem[M.~Bacastow et~al.(2019)M.~Bacastow, Van~Etten, and Weir]{offnadir}
M.~Bacastow, T., Van~Etten, A., and Weir, N.
\newblock {Automated Feature Extraction Using High Off-Nadir Satellite Imagery
  for Humanitarian Assistance and Disaster Response (HADR)}.
\newblock In \emph{GTC-DC 2019}, 2019.
\newblock URL \url{https://developer.nvidia.com/gtc-dc/2019/video/dc91263-vid}.

\bibitem[McInnes et~al.(2020)McInnes, Healy, and Melville]{mcinnes2020umap}
McInnes, L., Healy, J., and Melville, J.
\newblock Umap: Uniform manifold approximation and projection for dimension
  reduction, 2020.

\bibitem[Peng et~al.(2017)Peng, Usman, Kaushik, Hoffman, Wang, and
  Saenko]{peng2017visda}
Peng, X., Usman, B., Kaushik, N., Hoffman, J., Wang, D., and Saenko, K.
\newblock Visda: The visual domain adaptation challenge, 2017.

\bibitem[Richter et~al.(2016)Richter, Vineet, Roth, and
  Koltun]{richter2016playing}
Richter, S.~R., Vineet, V., Roth, S., and Koltun, V.
\newblock Playing for data: Ground truth from computer games, 2016.

\bibitem[Sainburg et~al.(2021)Sainburg, McInnes, and
  Gentner]{sainburg2021parametric}
Sainburg, T., McInnes, L., and Gentner, T.~Q.
\newblock Parametric umap embeddings for representation and semi-supervised
  learning, 2021.

\bibitem[Tsai et~al.(2020)Tsai, Hung, Schulter, Sohn, Yang, and
  Chandraker]{tsai2020learning}
Tsai, Y.-H., Hung, W.-C., Schulter, S., Sohn, K., Yang, M.-H., and Chandraker,
  M.
\newblock Learning to adapt structured output space for semantic segmentation,
  2020.

\bibitem[Vu et~al.(2019)Vu, Jain, Bucher, Cord, and Pérez]{vu2019advent}
Vu, T.-H., Jain, H., Bucher, M., Cord, M., and Pérez, P.
\newblock Advent: Adversarial entropy minimization for domain adaptation in
  semantic segmentation, 2019.

\bibitem[Xu et~al.(2019)Xu, Lu, Li, Khaitan, and Zaytseva]{xu2019building}
Xu, J.~Z., Lu, W., Li, Z., Khaitan, P., and Zaytseva, V.
\newblock Building damage detection in satellite imagery using convolutional
  neural networks, 2019.

\bibitem[Zbontar et~al.(2021)Zbontar, Jing, Misra, LeCun, and Deny]{barlow}
Zbontar, J., Jing, L., Misra, I., LeCun, Y., and Deny, S.
\newblock Barlow twins: Self-supervised learning via redundancy reduction,
  2021.

\bibitem[Zhang et~al.(2021)Zhang, Zhang, Zhang, Chen, Wang, and
  Wen]{zhang2021prototypical}
Zhang, P., Zhang, B., Zhang, T., Chen, D., Wang, Y., and Wen, F.
\newblock Prototypical pseudo label denoising and target structure learning for
  domain adaptive semantic segmentation, 2021.

\end{thebibliography}
\bibliographystyle{icml2021}

\appendix

\section{Effect of Augmentations on Domain Adaptation}
\label{appendix:a}

\begin{figure}
\centering     
\subfigure[Input]{\label{fig:vegas}\includegraphics[width=40mm]{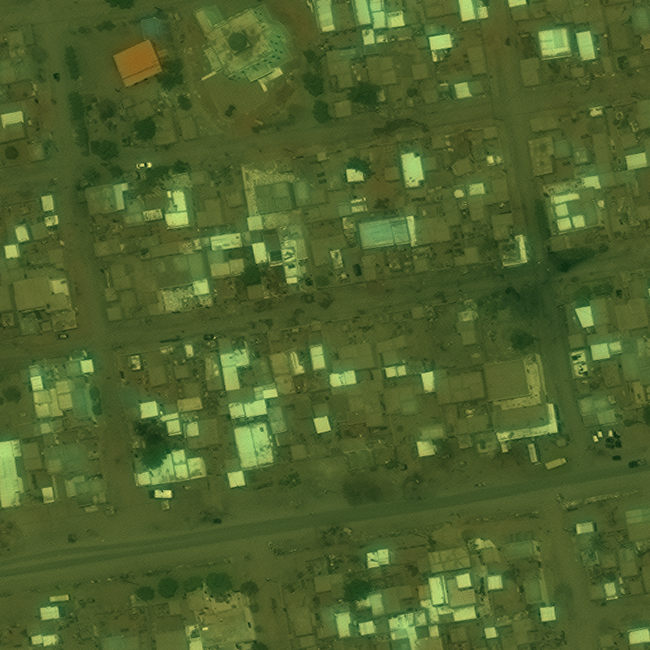}}
\subfigure[Ground Truth]{\label{fig:khartoum}\includegraphics[width=40mm]{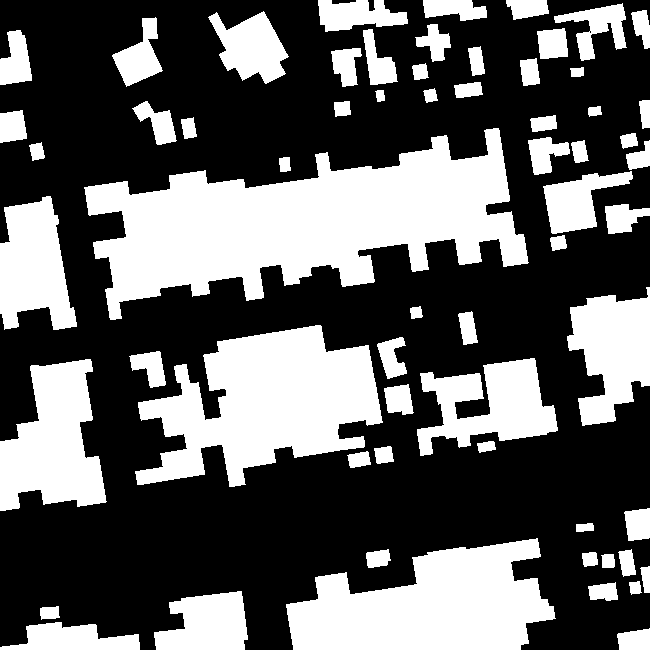}}
\subfigure[ADVENT]{\label{fig:onnadir}\includegraphics[width=40mm]{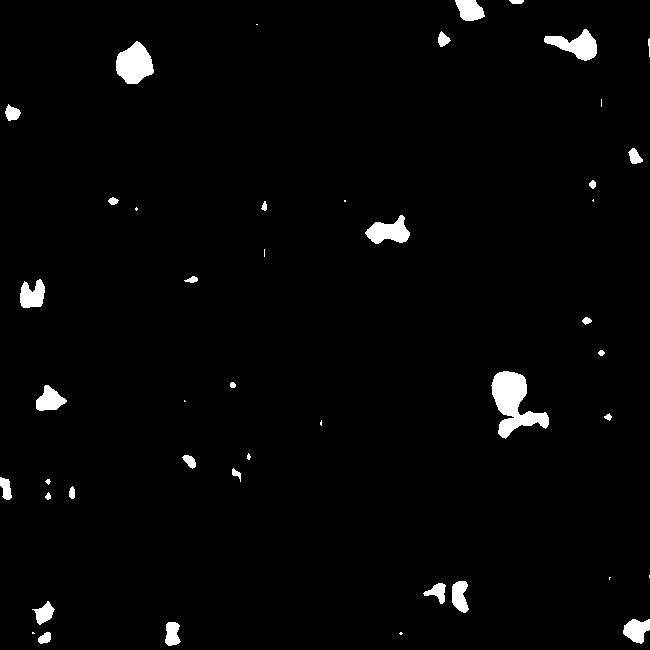}}
\subfigure[Augmented ADVENT]{\label{fig:voffnadir}\includegraphics[width=40mm]{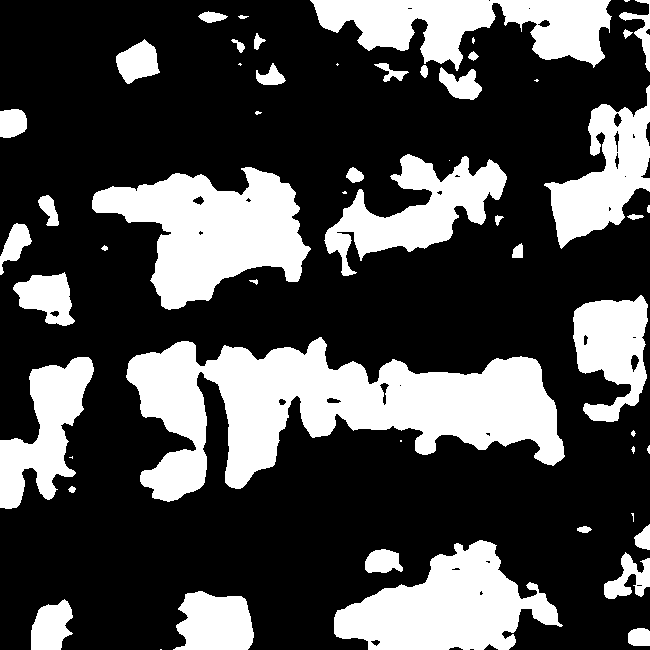}}
\caption{Example Vegas-to-Khartoum predictions for (a), from ADVENT, with and without augmentations (fixed per-augmentation probability of $0.6$).}
\label{newpreds}
\end{figure}

Early results using augmentations have been promising. For example, applying the augmentations of \citet{ada} with a fixed per-augmentation probability significantly improves both source-only and ADVENT performance, and improves ADVENT's performance \emph{relative} to source-only, as shown in Figure~\ref{newpreds} and Table~\ref{table2} for Vegas-to-Khartoum adaptation. It remains to be seen if this effect persists across different benchmarks for multiple runs and, if so, how it varies across per-augmentation probabilities.

\subsection{Effect of Augmentations on Adversarial and Source-Only Adaptation}

The observed improvement augmentation brings to both source-only and ADVENT-based adaptation is significantly greater than the \emph{relative} improvement it confers to ADVENT over source-only. As these augmentations are comparatively simple to apply, they represent a reasonable, unobtrusive starting-point for improving the domain adaptation of relevant supervised tasks in geospatial machine learning.

\subsection{Role of Adaptive Augmentation Probability}

Though we do not require adaptive augmentation probability for the reasons described by \citet{ada}, it remains of interest whether adaptive probability schedules can prove more useful during training than fixed probabilities.

\begin{table}[t]
\caption{Intersection over Union (IoU) for source-only and ADVENT training on Vegas-to-Khartoum adaptation, with and without augmentations (fixed per-augmentation probability of $0.6$).}
\label{table2}
\vskip 0.15in
\begin{center}
\begin{small}
\begin{sc}
\begin{tabular}{lccc}
\toprule
& Source-Only & ADVENT & $\Delta$ IoU \\
\midrule
No Aug.    & 15.09 &  13.59 & $-$1.50 \\
Aug.    & 33.81  & 36.00 & $+$2.19 \\
\bottomrule
\end{tabular}
\end{sc}
\end{small}
\end{center}
\vskip -0.1in
\end{table}


\end{document}